\documentclass[10pt,twocolumn,letterpaper]{article}

\usepackage{cvpr}              
\usepackage[accsupp]{axessibility}

\usepackage[dvipsnames]{xcolor}

\definecolor{cvprblue}{rgb}{0.21,0.49,0.74}
\usepackage[pagebackref,breaklinks,colorlinks,citecolor=cvprblue]{hyperref}

\def\paperID{48} 
\def\confName{CVPR}
\def\confYear{2024}

\title{PCQA: A Strong Baseline for AIGC Quality Assessment \\ 
Based on Prompt Condition}

\author{Xi Fang\footnotemark[1]\\
DP Technology\\
{\tt\small fangxi@dp.tech}
\and
Weigang Wang\footnotemark[1]\\
Cisco\\
{\tt\small weigwang@cisco.com}
\and
Xiaoxin Lv\footnotemark[1]\\
Shopee\\
{\tt\small xiaoxin.lv@shopee.com}
\and
Jun Yan\footnotemark[2]\\
TongJi University\\
{\tt\small yanjun@tongji.edu.cn}
}

\begin{document}
\maketitle

\renewcommand{\thefootnote}{\fnsymbol{footnote}} 
\footnotetext[1]{Xi Fang, Weigang Wang, and Xiaoxin Lv contributed equally to this work.} 
\footnotetext[2]{Jun Yan is corresponding author.} 

\begin{abstract}
The development of Large Language Models (LLM) and Diffusion Models brings the boom of Artificial Intelligence Generated Content (AIGC). It is essential to build an effective quality assessment framework to provide a quantifiable evaluation of different images or videos based on the AIGC technologies. The content generated by AIGC methods is driven by the crafted prompts. Therefore, it is intuitive that the prompts can also serve as the foundation of the AIGC quality assessment. This study proposes an effective AIGC quality assessment (QA) framework. First, we propose a hybrid prompt encoding method based on a dual-source CLIP (Contrastive Language-Image Pre-Training) text encoder to understand and respond to the prompt conditions. Second, we propose an ensemble-based feature mixer module to effectively blend the adapted prompt and vision features. The empirical study practices in two datasets: AIGIQA-20K (AI-Generated Image Quality Assessment database) and T2VQA-DB (Text-to-Video Quality Assessment DataBase), which validates the effectiveness of our proposed method: \textbf{P}rompt \textbf{C}ondition \textbf{Q}uality \textbf{A}ssessment (PCQA). Our proposed simple and feasible framework may promote research development in the multimodal generation field.
\end{abstract}    
\section{Introduction}
\label{sec:intro}

\begin{figure}[t]
    \centering
    \includegraphics[width=0.9\hsize]{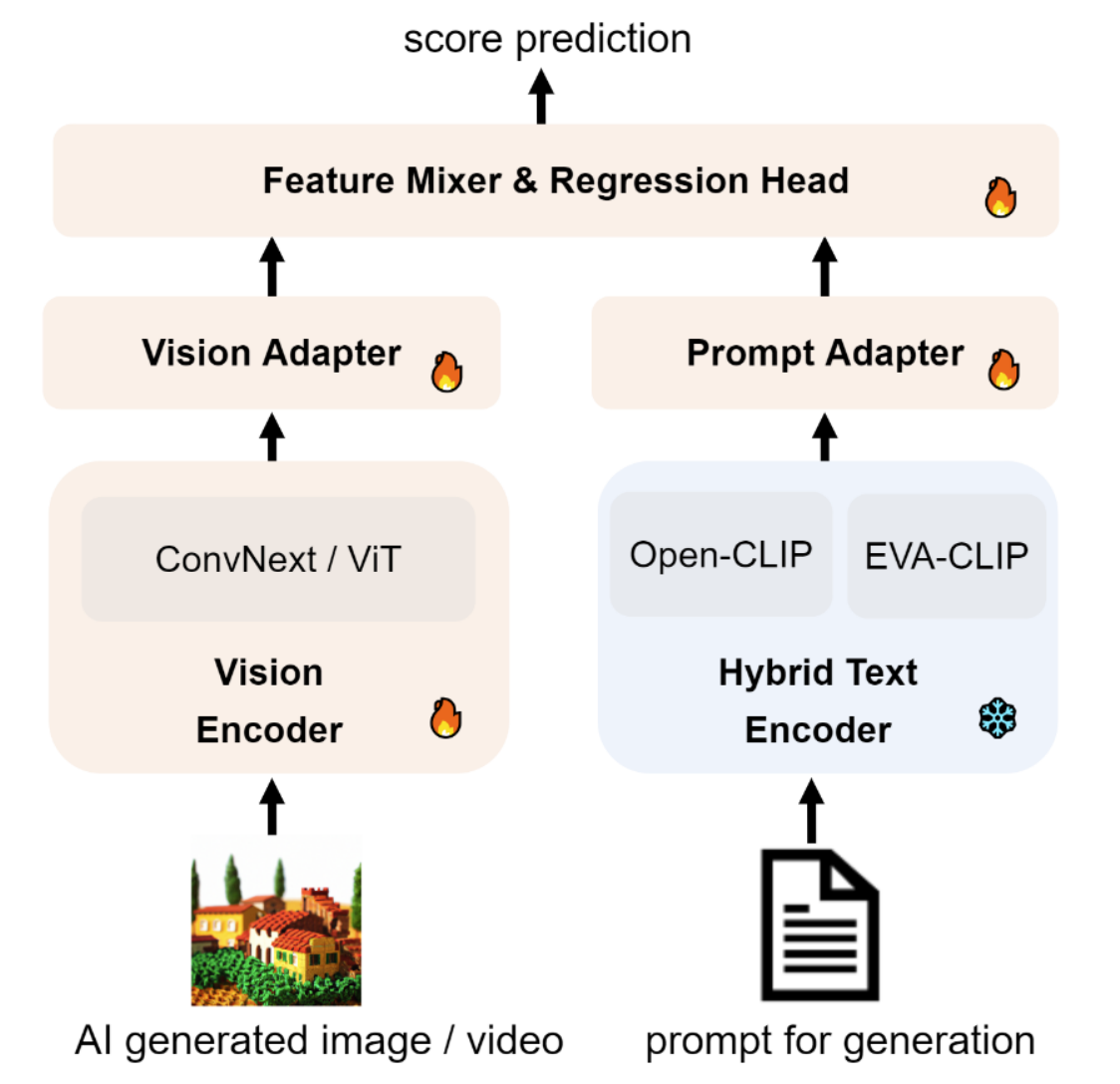}
    \caption{Overview of the \textbf{P}rompt \textbf{C}ondition \textbf{Q}uality \textbf{A}ssessment (PCQA) model. The AIGC content and the corresponding prompt used to generate it are input separately. The information from the prompt will be encoded by the hybrid CLIP text encoder and used as a condition for visual quality assessment, with the trainable feature adapter to align the feature from different modals. The final MOS regression result is obtained through a feature mixer and an MLP regressor.}
    \label{fig:arch}
\end{figure}

With the proliferation of Artificial Intelligence Generated Content (AIGC) technologies, AIGC images or videos have gradually appeared in people's view. The creation, sharing, and interaction of images and videos based on the AIGC technologies calls for their quality assessment. The Quality of Experience (QoE) guarantees that the AIGC works should align with the human aesthetic point of view and the lofty moral sense in artistic appreciation, like rejecting the vulgar stuff.
\par Currently, the quality assessment of User Generated Content (UGC) via deep neural networks is matured. The local binary pattern features based on the distortion aggravation mechanism can measure the similarities between the distorted image and the multiple pseudo reference images (MPRIs)~\cite{min2018blind}. The renaissance of deep learning brings a new paradigm of UGC quality assessment. The deep bilinear convolutional neural network (BCNN) can help the users implement the blind image quality assessment (BIQA)~\cite{zhang2018blind}. The end-to-end spatial feature extraction networks can directly learn quality-aware spatial feature representations of video frame pixels. The hierarchical feature fusion and iterative mixed database training can boost the QoE~\cite{sun2023blind}. The contrastive language image pre-training (CLIP) method~\cite{radford2021learning} builds the vision-language correspondence and forms a new multitask learning perspective in blind image quality assessment~\cite{zhang2023blind}. Meanwhile, the no-reference video quality assessment can be realized by the deep neural networks (DNNs) based on the content dependency and temporal-memory effects with the intuitions of human visual system~\cite{li2021unified}, multi-scale quality fusion strategy~\cite{sun2022deep}, or the disentangle of aesthetic perspectives and technical elements~\cite{wu2023exploring}. The CLIP method has also been scaled to the in-the-wild video quality assessment task~\cite{wu2023towards}.
\par In contrast to the assessment of User-Generated Content (UGC) quality, the evaluation of AI-Generated Content (AIGC) quality would place a greater emphasis on the high-level semantic information over the low-level details. The content produced by AI systems would exhibit a more profound alignment with the initial prompt, demonstrating enhanced coherence and relevance. This distinction underscores the AI's capability to synthesize and contextualize information, producing outputs that are not merely superficially related to the prompt but are intrinsically connected through higher-order semantic relationships.
\par Benjamin was one of the earliest thinkers to focus on the impact of technological advances, particularly the development of mechanical reproduction, on works of art~\cite{benjamin2018work}. The proliferation of technologies precipitates the erosion of the ``Aura", catalyzing the engagement of the wider public in both the creation and critique of art. As the societal value of art recedes, a growing chasm develops between critical and appreciative interactions from the audience. This divergence is becoming more pronounced in the epoch of AIGC, and machine evaluations inherit the same kind of subjectivity and bias that human evaluation has experienced.
\par It raises a worthwhile scientific problem: \textit{What kind of machine-learning-based assessment model can evaluate the artistic caliber of AIGC content with relative objectivity and less bias?}
\par In this study, we propose a unified framework for AIGC quality assessment based on the specified prompt condition denoted by Figure \ref{fig:arch}. It employs a dual-source CLIP text encoder (Open-Clip~\cite{radford2021learning, cherti2023reproducible} and EVA-CLIP~\cite{sun2023eva}) to interpret the prompts for the pair projections with the visual features extracted by the Vision Transformers (ViTs)~\cite{dosovitskiy2021image} and ConvNeXts~\cite{liu2022convnet}. Then, a feature mixer module blends the text and image features to construct the correlations between images/videos and assessment quality comments. Such a pipeline can drive the evaluation paradigm to focus on the high-level features of AIGC works. We adopt moderate training-time data augmentation to realize the trade-off between data diversity and aesthetic standards. The erasing of ``Aura" has caused subjectivity and bias in the aesthetic evaluation~\cite{benjamin2018work}. We design an ensemble method to mitigate the bias in the scoring process so that the decisions from the different vision backbones would be averaged. It mimics the scoring of multiple human reviewers at many art-judging events. The experimental results on AIGIQA-20K~\cite{lisncodalab2024imagecompetition} (AI-Generated Content Quality Assessment dataset) and T2VQA-DB (Text-to-Video Quality Assessment DataBase)~\cite{lisncodalab2024videocompetition} validate the effectiveness of our proposed method.
\par In summary, the major contributions of our work are listed as follows:
\begin{itemize}
    \item We propose a unified framework of AIGC image or video quality regression with the prompt condition, which focuses more on the high-level semantic information.
    \item We design the mechanism based on the feature adapter and feature mixer to enable effective interaction between the prompt condition and visual features.
    \item We propose a novel ensemble method to mitigate the bias in the quality assessment scoring process.
\end{itemize}
\par The organization of this manuscript is arranged as follows. Section 2 describes the related work in this research field. Section 3 provides a general view of our proposed method. Section 4 demonstrates the considerable experimental results of the proposed method in this study. Section 5 concludes the study and gives some further perspectives.

\section{Related Works}
\label{sec:related_works}
This section reviews the research development in UGC quality assessment, AIGC technologies, and mainstream multimodal learning methods in the past few years.

\subsection{UGC quality assessment}
\par The development of UGC quality assessment experiences three eras: the era before the proliferation of deep learning, the era with the utilization of deep learning technologies, and the era with the applications of multimodal learning.
\par The classical blind image/video quality assessment depends on the signal processing and classical machine learning methods, like wavelet transform~\cite{moorthy2011blind}, DCT transform~\cite{saad2012blind}, feature learning~\cite{ye2012unsupervised}, rank learning~\cite{ma2017dipiq}, and multiple pseudo reference image distortion aggregation~\cite{min2018blind}. These methods depend on the handcrafted feature engineering work.
\par The blooming of deep learning brings the new paradigm of UGC quality assessment. The milestone study utilizes shallow convolutional neural networks (CNNs) to implement the no-reference image quality assessment~\cite{kang2014convolutional}. The first image quality assessment network based on deep neural networks comprises ten convolutional layers and five pooling layers for feature extraction, and two fully connected layers for score computation~\cite{bosse2017deep}. At the same time, it is demonstrated that the distortion identification and quality prediction tasks can be jointly optimized in an end-to-end CNNs~\cite{ma2017end}, or handled by the bilinear convolutional neural network~\cite{zhang2018blind}. The influence of feature learning based on the pre-trained image classification task is also explored and exploited by the research community~\cite{bianco2018use}. Previous studies have proposed a hierarchical network to integrate the extracted features based on an iterative mixed database training strategy to realize the problem of quality-aware feature representation and insufficient training samples in terms of the content and distortion diversity~\cite{sun2023blind}. Recently, Wu et al.~\cite{wu2023assessor360} have proposed a multi-sequence network called Assessor360 for blind omnidirectional image quality assessment. The method achieves efficient assessment of omnidirectional image quality by designing Recursive Probabilistic Sampling (RPS) to generate viewport sequences, combining Multi-scale Feature Aggregation (MFA) and Distortion-Aware Blocks (DAB) for distortion and semantic features, as well as Temporal Sequence Modeling Module (TMM) for learning temporal variations of viewports. The no-reference video assessment problem is also significant. Li et al.~\cite{li2021unified} have investigated the problem of automating the quality assessment in in-the-wild videos and proposed a unified framework that improves the performance of video quality assessment models by combining the content-dependent and temporal memory effects of the human visual system and by employing a mixed dataset training strategy. Wang et al.~\cite{wang2021rich} provide an in-depth analysis of the correlation between video quality assessment model performance and video content, technical quality, and compression level. Sun et al.~\cite{sun2022deep} have proposed a simple but effective deep learning-based reference-free quality assessment model, which learns quality-aware spatial features of video frames through an end-to-end spatial feature extraction network and combines them with motion features to assess video quality, uses a multilayer perceptron (MLP) network for quality regression, and employs a multiscale quality fusion strategy to deal with the problem of assessing the quality of videos with different spatial resolutions. Wu et al.~\cite{wu2023exploring} propose a model called DOVER, which evaluates the quality of UGC videos from both aesthetic and technical perspectives and predicts the overall video quality through a fusion strategy with a subjective heuristic strategy.
\par CLIP~\cite{radford2021learning} is a multimodal pre-training technique that is trained on large-scale image and text datasets to achieve strong cross-modal comprehension and generalization. The novel technology has been applied to the UGC image~\cite{zhang2023blind} and video~\cite{wu2023towards} quality assessment. It inspires further exploration in the AIGC quality assessment task.

\subsection{Generative Models}
\par In the past decade, the generative models have demonstrated the surprising ability of content creation, including Generative Adversarial Networks (GANs)~\cite{goodfellow2014generative} and Variational Autoencoders~\cite{kingma2014auto}. 
\par The diffusion model is a neural network model based on the Markov decision process, which realizes the content creation with the multiple-step forward addition of noise and reverses operation of denoising~\cite{ho2020denoising,song2021score}. The Stable Diffusion model~\cite{rombach2022high} improves the image quality and the computation efficiency compared to the vanilla diffusion models~\cite{ho2020denoising}, and realizes the variety and consistency in image generation quality. Recently, the variant of the Stable Diffusion model realizes the generation of the images with the $1024 \times 1024$ resolution~\cite{podell2024sdxl}. Surprisingly, the stable latent diffusion models can also be applied to create the video content~\cite{blattmann2023stable}. Peebles et al.~\cite{peebles2023scalable} have explored a new class of diffusion models based on transformer architectures~\cite{vaswani2017attention,dosovitskiy2021image}, which boosts the feature extraction and representation ability of the diffusion models. The diffusion transformer (DiT) model inspires the surprising Sora Large Vision Model (LVM)~\cite{openai2024sora}. DreamBooth~\cite{ruiz2023dreambooth} is an extension of the text-to-image diffusion models that allows for fine-tuning the specific prompts. The DALL-E models are also the variants of diffusion models, which can generate high-quality images with the guidance of CLIP method~\cite{ramesh2022hierarchical} or Large Language Models (LLMs)~\cite{betker2023improving}. Based on the proliferation of AIGC technologies, some benchmarks have also been released. The AGIQA-3K is an open database for the generative image quality assessment~\cite{li2023agiqa}. Recently, two larger datasets of image~\cite{lisncodalab2024imagecompetition} and video quality~\cite{lisncodalab2024videocompetition} assessment have been released. It lays a cornerstone for the subsequent research on related evaluation methods.

\subsection{Contrastive Language-image Pre-training}
\par One of the beautiful wishes of computer vision is that machine vision can operate like human vision. The research community has made huge efforts to learn visual representations that correspond with the semantic information. The CLIP method~\cite{radford2021learning} is trained using large-scale image and text pairs, which can be natural language descriptions, labels, or other forms of annotations. During training, the model is optimized to ensure that a specific language signal is close to its corresponding image in the feature space while leaving itself in the feature space for mismatched image and text pairs. The SLIP (Self-supervised Language-Image Pre-training) method~\cite{mu2022slip} introduces the auxiliary enhancement of feature representation via the self-learning paradigm. The BLIP (Language-image Pre-training) methods~\cite{li2022blip,li2023blip} focus on learning the complex relationships between images and text through bidirectional reconstruction. Recently, Yang et al.~\cite{yang2023attentive} propose an attentive token removal approach based on the random mask to accelerate the training time of CLIP. Overall, this field is just starting and will exert a methodological influence on the AIGC quality assessment task.


\section{Proposed Approach}
\label{sec:method}

We propose a prompt-conditional quality assessment method, which can be utilized for both AIGC image and video quality assessment tasks. First, our method encodes the image features and prompts text features separately, using trainable image encoder and frozen CLIP text encoder. Subsequently, the text features are employed as conditions to interact with the image features, culminating in the regression of the Mean Opinion Score (MOS).

\subsection{Quality Assessment with Prompt Condition}
\par In the domain of quality assessment, traditional approaches have separated image and video QA tasks into distinct categories and only take a single image or single video as input. For image QA, the score is determined solely based on the quality of the image input, denoted as Eq. (\ref{ymosiqa}). 
\begin{equation}\label{ymosiqa}
y_{\operatorname{mos}}=f_{I Q A}(x)
\end{equation}
where the symbol $ \hat{y_{mos}}$ stands for the image quality assessment (IQA) on the data $x$. Similarly, video QA tasks utilize a separate scoring function, given by the formulation defined in Eq. (\ref{ymosvqa})
\begin{equation}\label{ymosvqa}
y_{\operatorname{mos}}=f_{V Q A}(x)
\end{equation}
AIGC content is accompanied by the prompt text that generates it, and assessing the quality of AIGC content must take into account the alignment between these contents and their corresponding prompts. Traditional methods, however, overlook this aspect. Thus, we proposed the \textbf{P}rompt \textbf{C}onditional \textbf{Q}uality \textbf{A}ssessment (PCQA) method, which uses prompt as a condition to assess the AIGC quality, denoted as Eq. (\ref{pcqa}), where $x$ is the AIGC video input and $t$ is the prompt text input as a condition. When the video has only one frame, this method degenerates into the image quality assessment.

\begin{equation}
\label{pcqa}
\hat{y_{mos}}=f_{PCQA}(x | t)
\end{equation}

The whole framework of our proposed PCQA method has been demonstrated in Figure \ref{fig:arch}. The network architecture comprises a trainable visual encoder, a frozen hybrid text encoder, as well as trainable feature adapters, feature mixers, and regression heads.
We simultaneously input AIGC images or videos, along with the corresponding prompt texts that are used to generate this content and regard the prompt texts as conditions for the Mean Opinion Score (MOS) regression.

\subsection{Hybrid Text Encoder}
\par CLIP~\cite{radford2021learning} model is pretrained on large number of image-text pairs. It enables the cross-modal understanding between language and images. In particular, it can guide image or video generation and editing through natural language, as known as ``prompt'', which opens up the possibility of creating and understanding new works of art. Therefore, it inspires us to encode our prompt information in AIGC quality assessment task based on the CLIP mechanism. 
\par However, the CLIP text encoder is typically challenging to finetune. Too many trainable parameters in the model make training more hardware demanding and hyper-parameters tuning more difficult. So we froze the parameters of CLIP text encoder during training and added a trainable feature adapter which enabled the output features to better adapt to the task. Freezing the text encoder during training can make the entire model more amenable to training. This is also proven in the experiment. For further details, refer to Section \ref{exp}.
\par The text encoders used in AIGC methods originate from various sources. We have integrated multiple CLIP text encoders and concatenated their outputs to enhance the content of the extracted textual information. Utilizing the frozen CLIP text encoder, we encode prompts from two distinct open-source implementations: Open-CLIP\cite{cherti2023reproducible} and EVA-CLIP\cite{sun2023eva}, which pretrained on diverse datasets such as DFN-5B\cite{fang2023data}, LAION-5B\cite{schuhmann2022laion}, DataComp-1B\cite{gadre2024datacomp}, and WebLI\cite{chen2022pali}. This design enhances the information from text condition. 

\begin{figure}[!t]
    \centering
    \includegraphics[width=0.65\hsize]{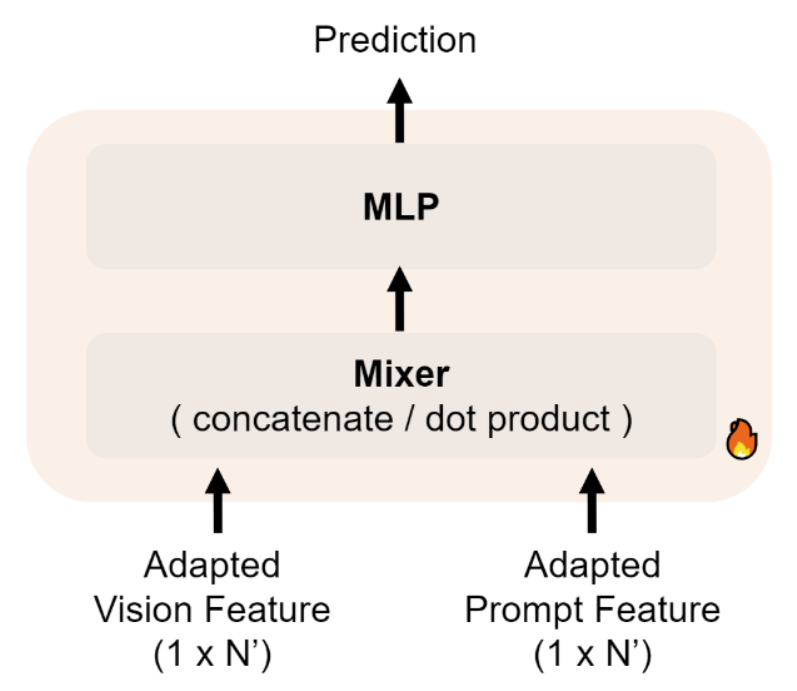}
    \caption{Feature mixer and regression head. Concatenation or dot product are used as feature mixer. This enables the visual features and the textual features of the prompt to interact.}
    \label{fig:mixer}
\end{figure}

\subsection{Feature Adapter and Mixer}
\par We introduce a trainable dense layer that functions as a prompt adapter to enhance the synergy between textual and visual elements. For the visual aspect, we utilize a trainable vision backbone equipped with ImageNet pre-trained weights to extract the visual features, with ConvNeXt-Small~\cite{liu2022convnet} serving as the standard choice. This visual backbone, characterized by its modern architecture and extensive receptive field, is adept at extracting high-level semantic information from images. Regarding video input, we methodically extract visual features from a maximum of 16 frames and integrate these features by applying a 1D Convolutional Neural Network supplemented by mean pooling. After extracting the visual features, a trainable dense layer is employed as a vision adapter. Similarly, the concatenated textual features from CLIP are processed through a trainable dense layer designated as a feature adapter for prompt information.
\par We then integrate a feature mixer module, as shown in Figure \ref{fig:mixer}, which employs both the dot product and concatenation techniques to foster a compelling interplay between the adapted prompt and vision features, akin to the cross-attention in transformers. The dot product mixer excels at capturing the correlation between the generated images and the prompts, while the concatenation treats the prompt as a conditional factor. These mixers are applied by various experts and contribute to the model blending.
\par Finally, the merged features are fused by a two-layer MLP to predict the ultimate quality score. This approach ensures a nuanced and comprehensive quality assessment sensitive to the alignment between the AIGC image or video and its generating prompts.

\subsection{Ensemble Method in Quality Assessment}
\begin{figure}
    \centering
    \includegraphics[width=0.85\hsize]{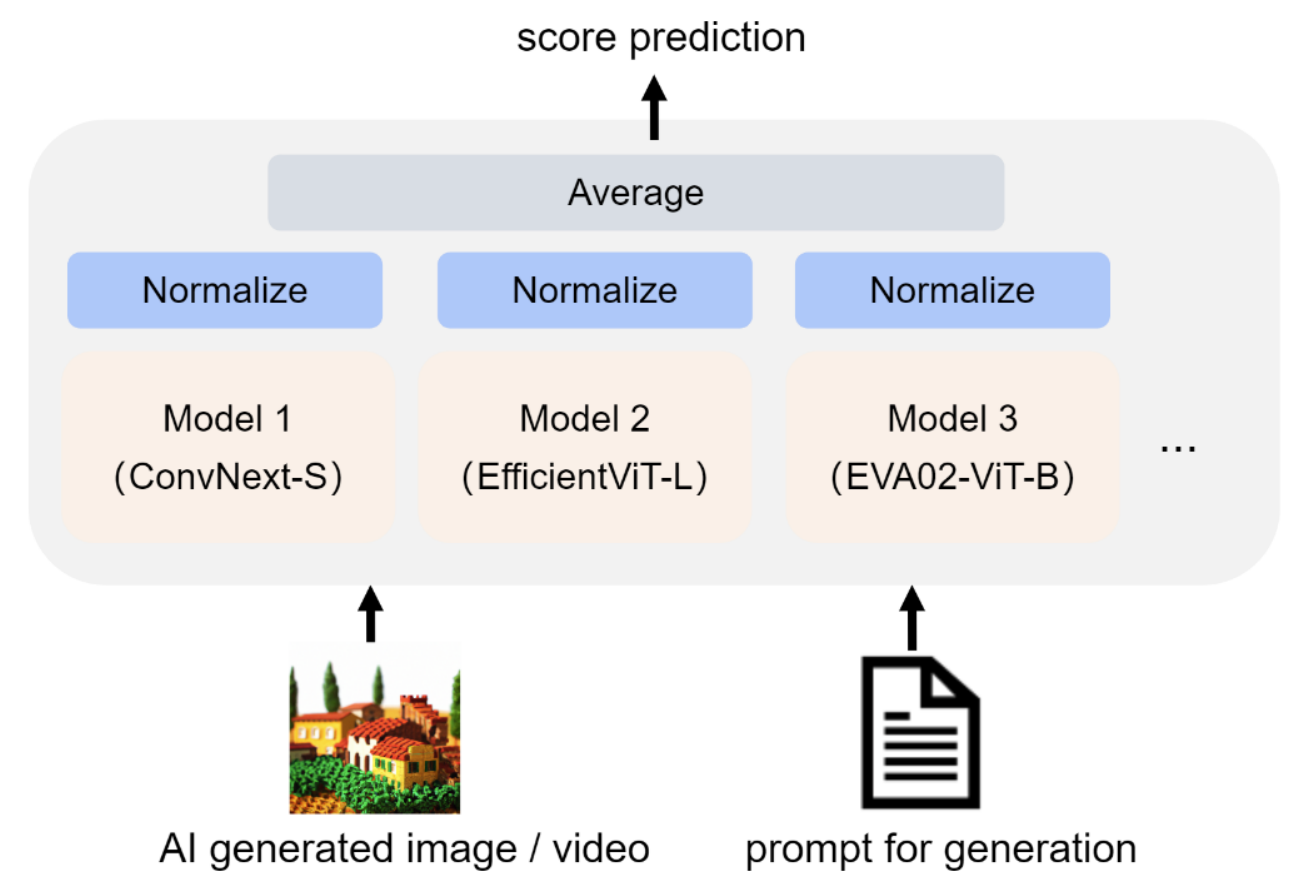}
    \caption{Overview of the final quality score computation strategy by model ensemble. The final score is average blending 3 models with different vision backbone.}
    \label{fig:ensemble}
\end{figure}

We ensemble three different models with different vision backbones: ConvNeXt-S, EfficientVit-L, and EVA02-Transformer-B. Firstly, we normalize the predicted scores of each model on the test dataset, so that different models have the same mean value and variance of prediction. After the normalization operation, we blend all the models by averaging the MOS prediction. Figure \ref{fig:ensemble} shows the pipeline of our ensemble method for quality assessment.

We adopt a strategy of normalizing the output of each individual model before average blending. By combining multiple models, this ensemble method can reduce the prediction variance, thus reducing the overall generalization error. This ensemble approach can also prevent biases in the prediction of MOS scores across different models, ensuring that each model contributes equally to the final predicted score. Through this normalization process, we can balance the influence of each model, making their roles in the ensemble prediction more fair and effective. This not only enhances the accuracy of the prediction but also strengthens the robustness of the ensemble modeling method. More imaginatively, we can think of this ensemble approach as different experts in the assessment of human art. The individual evaluations may be potentially subjective, but the integration of multiple experts removes the variance.
\par We calculate the mean of normalized predicted values from multiple models as the final prediction of the MOS value. Eq. (\ref{blending}) formulates the mechanism. The symbol $x$ denotes the input of an image or video, and the symbol $t$ denotes the input prompt text. For the model $f_{i}$ in our normalized average blending method, $\mu_{i}$, $\sigma_{i}$ are the mean and variance for the prediction in the testing dataset, respectively. 
\begin{equation}
\label{blending}
\hat{y_{mos}} = \text{E}_{i}(\frac{f_{i}(x | t) - \mu_{i}}{\sigma_{i}}) 
\end{equation}


\section{Experimental Results}
\label{sec:experiment}

\subsection{Datasets}
\par During the NTIRE 2024 Competition, two novel datasets are introduced for the assessment of AI-generated content (AIGC) quality. Track 1 of the NTIRE Quality Assessment for AI-Generated Content Competition presents a benchmark dataset AIGIQA-20K~\cite{lisncodalab2024imagecompetition} designed for the evaluation of image quality. Concurrently, Track 2 delivers a benchmark dataset T2VQA-DB~\cite{lisncodalab2024videocompetition} tailored for the quality assessment of video content. These datasets contribute significantly to the accurate prediction of quality in AI-generated images and videos, thereby laying a crucial benchmark for advancements in multimodal learning methodologies.
\par AIGIQA-20K~\cite{lisncodalab2024imagecompetition} represents a new dataset featuring a broad array of content pertaining to art creation. It encompasses a training corpus of 14,000 images generated through the use of textual prompts. The primary predictive label employed is the mean opinion score (MOS). For evaluative purposes, the dataset is divided into a validation set, which consists of 2,000 samples and is used for Leaderboard-A rankings, and a test set, comprising 4,000 samples with their corresponding prompts, which is designated for Leaderboard-B assessments.
\par T2VQA-DB~\cite{lisncodalab2024videocompetition} offers an extensive collection designed for text-to-video generation research. It comprises 7,000 training videos, each accompanied by a textual prompt, and evaluated using mean opinion score (MOS) metrics. Additionally, the dataset includes 1,000 validation videos, complete with prompts for preliminary assessment on the Leaderboard-A, and a set of 2,000 test videos, also with associated prompts, for the evaluation on the Leaderboard-B.

\subsection{Implement Details}
\par The backbone of the proposed framework is ConvNeXt-Small \cite{liu2022convnet}, which serves as the feature extractor for our vision encoder. To enhance the model's performance, we incorporate EfficientVit-Large \cite{cai2023efficientvit} and EVA-02 \cite{fang2023eva} model to form a hybrid network. These models contribute to a strong ensemble effect.
\par For the construction of the hybrid text encoder, we employ the ``ViT-H-14-quickgelu-dfn5b" parameters sourced from the Open-CLIP model \cite{cherti2023reproducible}, which are pre-trained on the DataComp-1B dataset \cite{fang2023data}, as well as adopting the EVA-CLIP \cite{sun2023eva} weights. These weights play a crucial role in the encoding of prompt features and remain unchanged during our training phase to preserve the integrity of their pre-learned representations.
\par To seamlessly integrate visual and textual data into a unified space, we employ the vision and prompt adapters that leverage a dense layer with a 1024-dimensional latent space. When processing video inputs, we apply a two-layer convolutional neural network with a kernel size of three, followed by a pooling layer to extract relevant features. Additionally, a multi-layer perceptron (MLP) serves as the regression head to refine the model's predictions.
\par The training process involves 50 epochs with the AdamW optimizer~\cite{loshchilov2019decoupled}, which uses a weight decay of $1\times 10^{-2}$ and a learning rate of $2\times 10^{-5}$. We also implement a cosine learning rate decay, a warm-up strategy, auto-mixed-precision training, and a gradient clipping method with a normalized value of 1.0 to ensure the stable and efficient optimization. All the experiments are done with only one NVIDIA V100 card.

Throughout the training process, we modulate the input resolution from $448 \times 640$ pixels, striking a balance between the computational demand and the model efficacy with a consistent batch size of 16. To enhance the model robustness, we implement the data augmentation techniques, such as random horizontal flips, slight random resized crops, and subtle brightness and contrast adjustments, which are designed to be non-intrusive and maintain the images' perceptual quality.

The training objective is the reduction of mean squared error (MSE) between the predicted outputs and the normalized Mean Opinion Score (MOS), eschewing the use of ancillary external datasets. Employing normalized MOS as our regression target substantially bolsters the stability of the model throughout its training.

\subsection{Main Results}
\par SRCC (Spearman's Rank Correlation Coefficient) and PLCC (Pearson Linear Correlation Coefficient) represent the performance metrics within the validation dataset. Val Score is the mean value of SRCC and PLCC in the validation dataset (Leaderboard-A). The Test Score, on the other hand, refers to the competition's final score on the test dataset. The calculation method of the Test Score is the mean of SRCC and PLCC on the testing set (Leaderboard-B). 

\begin{table}[!t]
\centering
\caption{Competition Results in AIGIQA-20K}
\label{tab:score_comparison_image}
\begin{tabular}{@{}lcccc@{}}
\toprule
Method & SRCC & PLCC & Val Score & Test Score \\
\midrule
StairIQA\cite{sun2023blind} & 0.61 & 0.65 & 0.63 & 0.62 \\
\textbf{Ours} & \textbf{0.90} & \textbf{0.93} & \textbf{0.92} & \textbf{0.92} \\
\bottomrule
\end{tabular}
\end{table}

\begin{table}[!t]
\centering
\caption{Competition Results in T2VQA-DB}
\label{tab:score_comparison_video}
\begin{tabular}{@{}lcccc@{}}
\toprule
Method & SRCC & PLCC & Val Score & Test Score \\
\midrule
SimpleVQA\cite{sun2022deep} & 0.65 & 0.67 & 0.66 & 0.65 \\
\textbf{Ours} & \textbf{0.82} & \textbf{0.84} & \textbf{0.83} & \textbf{0.82} \\
\bottomrule
\end{tabular}
\end{table}

Table \ref{tab:score_comparison_image} and \ref{tab:score_comparison_video} demonstrate the considerable performance of the PCQA method in the AIGC image and video assessment. It is notable that the proposed method significantly surpasses the baseline method \cite{sun2023blind} \cite{sun2022deep}.

\subsection{Ablation Studies}
\label{exp}
\begin{figure}[!t]
    \centering
    \includegraphics[width=0.9\hsize]{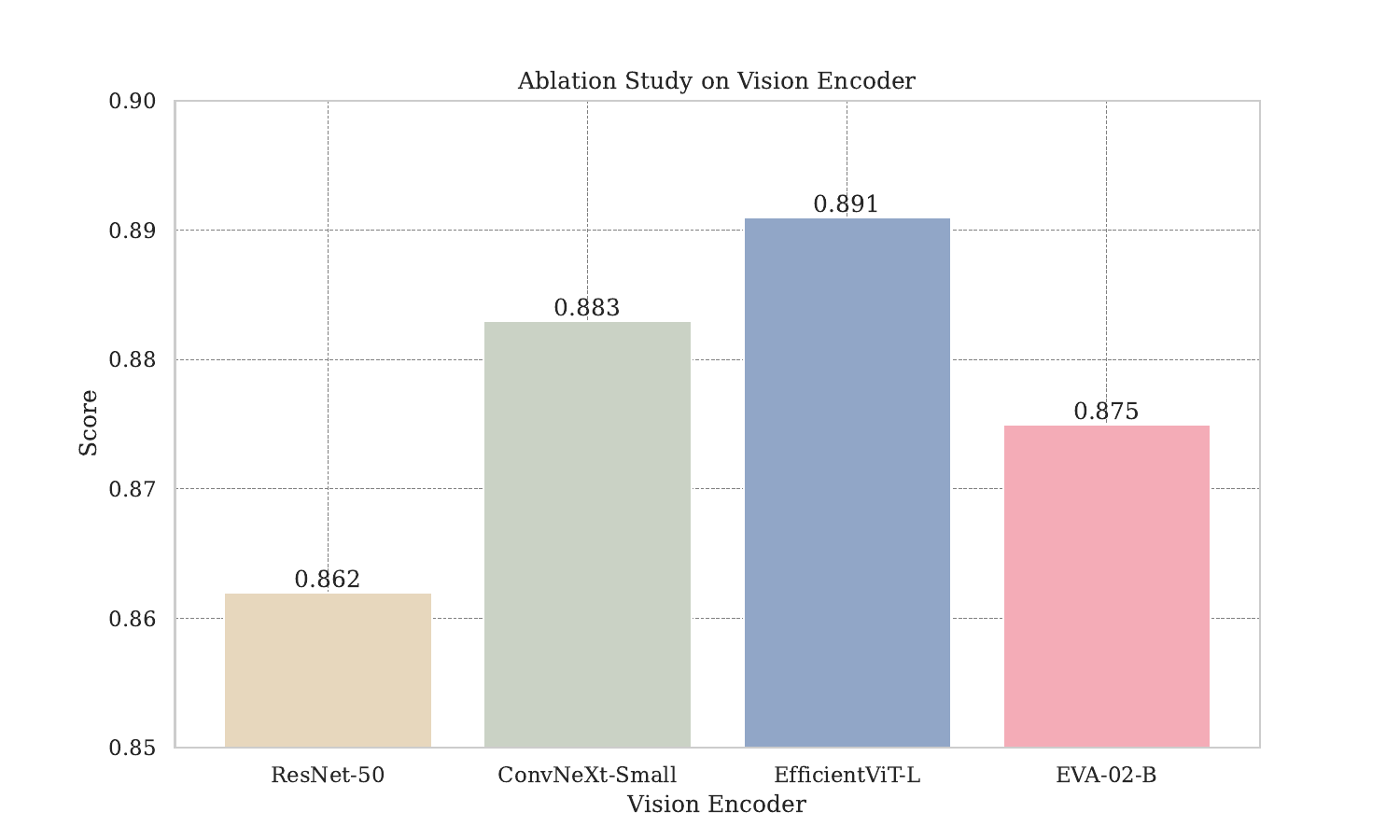}
    \caption{Ablation study on vision encoder choice. The score represents the average of SRCC and PLCC obtained through cross-validation on the AIGIQA-20K dataset.}
    \label{fig:backbone}
\end{figure}
To validate the effectiveness of the selection strategy of the image encoders, the text encoders, the design of feature mixers, and the model integration, we have devised a series of ablation study experiments. The score denotes the mean value of PLCC and SRCC on the five-fold cross-validation experiment.

\textbf{Vision Encoder:} We have explored various vision backbones and input resolution strategies through the empirical study. As shown in Figure  \ref{fig:backbone}, we have found that compared to the traditional design paradigm visual backbone like ResNet-50 \cite{he2016deep}, novel network architectures, such as ConvNeXt\cite{liu2022convnet} and variants of ViT\cite{dosovitskiy2020image, cai2023efficientvit, sun2023eva}, achieve significantly better performance. Under similar model sizes and inference latency, those networks that are designed with larger receptive fields and possess enhanced capabilities for high-level feature extraction yield better performance.

\begin{figure}
    \centering
    \includegraphics[width=0.8\linewidth]{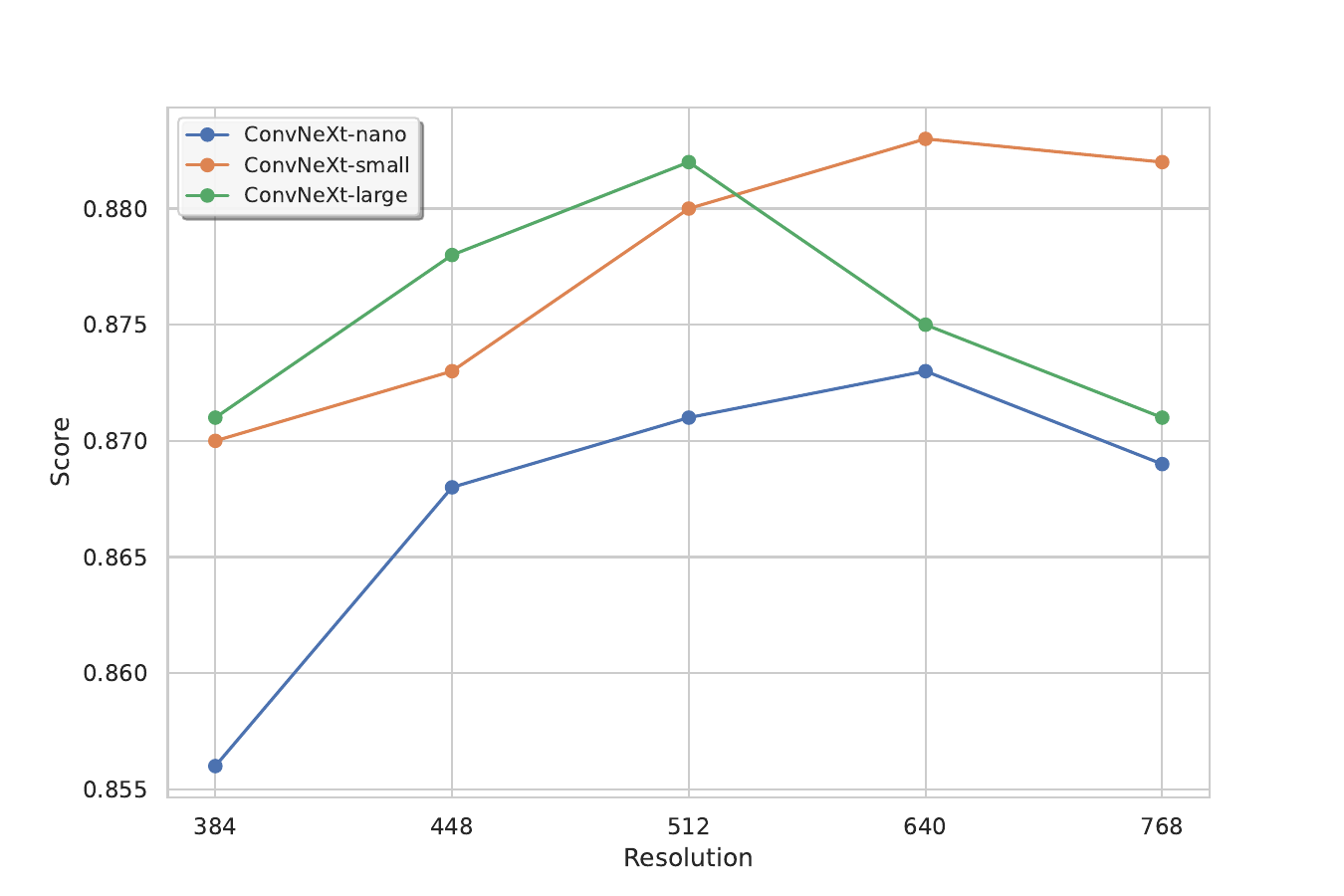}
    \caption{Ablation study on input resolution and model size. Input resolution between 448 to 640 leads to better results. Models with medium or larger sizes are also more likely to achieve better results.}
    \label{fig:resolution}
\end{figure}

We also explore the impact of input resolution and model size. The result is shown in Figure  \ref{fig:resolution}. A medium-sized model can achieve relatively better results at higher input resolutions.

\textbf{Text Encoder:} We compare the pretrained text encoders implemented with different CLIP approaches and found that the scheme utilizing a frozen pair of text encoders yielded the best results. The use of more than two text encoders does not lead to additional improvements. Additionally, we observe that many prompt texts are lengthy, which inspires us to explore the utilization of the text encoder based on Long-CLIP~\cite{zhang2024long}. However, it does not result in any performance enhancement. The details are displayed in Table \ref{tab:text}.

\begin{table}[!t]
\centering
\caption{Ablation Study on Text Encoder}
\label{tab:text}
\begin{tabular}{@{}lcc@{}}
\toprule
Text Encoder & Trainable & AIGIQA-20K \\
\midrule
EVA-CLIP\cite{sun2023eva} & \checkmark & 0.680 \\
EVA-CLIP\cite{sun2023eva} & $\times$ &  0.902 \\
Open-CLIP\cite{cherti2023reproducible} & $\times$ &  0.903 \\
Long-CLIP\cite{zhang2024long}  & $\times$ &  0.901 \\
Hybrid 2 Text Encoder & $\times$ &  \textbf{0.905} \\
Hybrid 3 Text Encoder & $\times$ &  \textbf{0.905} \\
\bottomrule
\end{tabular}
\end{table}

\textbf{Feature Mixer:} We have conducted the experiments with two distinct feature mixing approaches, precisely the operation of concatenation and dot product, and have observed no significant performance disparity between them. Consequently, we randomly select one of these methods for model construction and utilize it in the process of model fusion to enhance diversity.

\begin{table}[!t]
\centering
\caption{Ablation Study on Feature Mixer}
\label{tab:mixer}
\begin{tabular}{@{}lcc@{}}
\toprule
Feature Mixer & AIGIQA-20K & T2VQA-DB \\
\midrule
concatenation & 0.898 & \textbf{0.799} \\
dot product & \textbf{0.903} & 0.795 \\
\bottomrule
\end{tabular}
\end{table}

\textbf{Model Ensemble:} We explore the efficacy of ensemble methods in enhancing the robustness of models. Specifically, we have conducted experiments involving horizontal flipping of images in the test set as a form of test-time augmentation. Additionally, we explore the model ensemble mechanism with different visual backbones. Our ensemble method utilizes mean blending, which involves averaging the outputs of the models. In the image track, we ensemble three models, while in the video track, we ensemble two models.

\begin{table}[!t]
\centering
\caption{Ablation Study on Model Ensemble in AIGIQA-20K}
\label{tab:ensemble_i}
\begin{tabular}{@{}ll@{}}
\toprule
Backbone & Val Score \\
\midrule
ConvNeXt-Small & 0.898 \\
+ TTA & 0.911 (+0.013) \\
+ Ensemble EfficientViT-L & 0.915 (+0.017) \\
+ Ensemble EVA-02 & \textbf{0.916 (+0.018)} \\
\bottomrule
\end{tabular}
\end{table}

\begin{table}[!t]
\centering
\caption{Ablation Study on Model Ensemble in T2VQA-DB}
\label{tab:ensemble_v}
\begin{tabular}{@{}lc@{}}
\toprule
Backbone & Val Score \\
\midrule
ConvNeXt-Small & 0.799 \\
EfficientViT-L & 0.803 \\
Ensemble both & \textbf{0.815} \\
\bottomrule
\end{tabular}
\end{table}

We can observe that the model ensemble method through mean blending has led to improvements in both two track tasks (AIGC image QA and AIGC video QA). The results indicate that employing an ensemble learning approach is an extremely effective strategy for the AIGC-QA task.

\subsection{Discussion}
\subsubsection{Aspect Ratio Preservation}
Our approach involves directly resizing the original images to achieve computational efficiency. Such a mechanism compromises the aesthetic information inherent in the native aspect ratio. This methodological choice may lead to a loss of critical visual cues that contribute to the overall quality assessment. Future work should explore alternative pre-processing techniques that preserve the aspect ratio while maintaining computational efficiency, such as adaptive resizing or aspect ratio-aware cropping, to ensure a more interpretable representation of the image's visual content.

\subsubsection{Spatial Information Preservation}
In an effort to adapt our model to both AIGC image and video quality assessment tasks, we utilize the embeddings extracted at the backbone stage 4 (after global average pooling) from the visual backbone. While streamlining the model and significantly improving inference speed, this decision leads to a loss of spatial information that could potentially diminish performance in video quality evaluation. Future research should consider incorporating additional modules or techniques that capture spatial-temporal information to enhance the model's capability in video quality assessment.

\subsubsection{Length Extrapolation for AIGC-Video QA}
The T2VQA-DB Dataset~\cite{lisncodalab2024videocompetition} is composed of videos that are uniformly sampled to include 16 frames, whereas the AIGIQA-20K Dataset~\cite{lisncodalab2024imagecompetition} can be regarded as a variant of the T2VQA-DB dataset, restricted to a single frame. Our methodology has not been evaluated on video samples encompassing more extensive frame sequences. This restriction could impede the model's generalizability to real-world contexts, where the lengths of videos and frame rates are subject to significant variation. It is imperative for future research to assess the model's performance on datasets characterized by a wide array of frame counts and temporal resolutions, thereby confirming its suitability for a more expansive spectrum of video content.

\subsubsection{Computational Costs in Ensemble Method}
The ensemble methodology implemented in our study indeed augments the robustness of our model's predictions, along with the trade-off of the latency during inference and escalated training costs. To mitigate the increase in inference latency, we explore the self-distillation techniques~\cite{fang2021stochastic} that aim to distill the predictions from an ensemble of models into a more compact model. Nonetheless, this strategy incurs a slight diminution in performance. Future investigations should delve into more sophisticated ensembling techniques that reconcile computational efficiency with the precision of the model. It could be achieved through the employment of model compression methods or the optimization of ensemble learning algorithms.

\subsubsection{Competition Results}
We construct a universal strong baseline for the AIGC image and video quality assessment tasks through the design of a hybrid text encoder and feature adapter alongside an ensemble method utilizing multiple visual backbones combined with the mild data augmentation strategy. This framework achieves notable success, occupying a top-three position in the image track and a top-four ranking in the video track at the NTIRE competition~\cite{liu2024challenge}.

\section{Conclusion}
\par This paper proposes a unified quality assessment framework to provide the aesthetic evaluation for the AIGC images and videos. The incorporation framework of a hybrid CLIP text encoder and ensemble feature representation demonstrates its effectiveness in the empirical study. Our work can serve as a reference for some future AIGC artwork evaluation studies.
\par However, there are several remaining issues. First of all, the current method cannot adapt itself to the input image or video aspect ratio. Second, our proposed ensemble framework obtains a gain in performance with three times the inference time. Third, the prompt mechanism depends on the high-level semantic information and puts aside the concern about the low-level corruption. Last but not least, all the video datasets are pumped with only sixteen frames, and the video lengths are short. Therefore, our method needs to be tested on longer videos to verify its validity.
\par In the future, we will make our method adapt to different sizes instead of directly resizing the images. Moreover, reducing the inference time of the ensemble method and parsing the semantics of distortion in AIGC works would be significant. Finally, mainstream large models like Sora~\cite{openai2024sora} can generate videos in around one minute. Therefore, it is significant to evaluate our proposed method in long-length generative videos.

{
    \small
    \bibliographystyle{unsrt}
    \bibliography{main}

\begin{thebibliography}{10}

\bibitem{min2018blind}
Xiongkuo Min, Guangtao Zhai, Ke~Gu, Yutao Liu, and Xiaokang Yang.
\newblock Blind image quality estimation via distortion aggravation.
\newblock {\em IEEE Transactions on Broadcasting}, 64(2):508--517, 2018.

\bibitem{zhang2018blind}
Weixia Zhang, Kede Ma, Jia Yan, Dexiang Deng, and Zhou Wang.
\newblock Blind image quality assessment using a deep bilinear convolutional neural network.
\newblock {\em IEEE Transactions on Circuits and Systems for Video Technology}, 30(1):36--47, 2018.

\bibitem{sun2023blind}
Wei Sun, Xiongkuo Min, Danyang Tu, Siwei Ma, and Guangtao Zhai.
\newblock Blind quality assessment for in-the-wild images via hierarchical feature fusion and iterative mixed database training.
\newblock {\em IEEE Journal of Selected Topics in Signal Processing}, 2023.

\bibitem{radford2021learning}
Alec Radford, Jong~Wook Kim, Chris Hallacy, Aditya Ramesh, Gabriel Goh, Sandhini Agarwal, Girish Sastry, Amanda Askell, Pamela Mishkin, Jack Clark, et~al.
\newblock Learning transferable visual models from natural language supervision.
\newblock In {\em International conference on machine learning}, pages 8748--8763. PMLR, 2021.

\bibitem{zhang2023blind}
Weixia Zhang, Guangtao Zhai, Ying Wei, Xiaokang Yang, and Kede Ma.
\newblock Blind image quality assessment via vision-language correspondence: A multitask learning perspective.
\newblock In {\em Proceedings of the IEEE/CVF conference on computer vision and pattern recognition}, pages 14071--14081, 2023.

\bibitem{li2021unified}
Dingquan Li, Tingting Jiang, and Ming Jiang.
\newblock Unified quality assessment of in-the-wild videos with mixed datasets training.
\newblock {\em International Journal of Computer Vision}, 129(4):1238--1257, 2021.

\bibitem{sun2022deep}
Wei Sun, Xiongkuo Min, Wei Lu, and Guangtao Zhai.
\newblock A deep learning based no-reference quality assessment model for ugc videos.
\newblock In {\em Proceedings of the 30th ACM International Conference on Multimedia}, pages 856--865, 2022.

\bibitem{wu2023exploring}
Haoning Wu, Erli Zhang, Liang Liao, Chaofeng Chen, Jingwen Hou, Annan Wang, Wenxiu Sun, Qiong Yan, and Weisi Lin.
\newblock Exploring video quality assessment on user generated contents from aesthetic and technical perspectives.
\newblock In {\em Proceedings of the IEEE/CVF International Conference on Computer Vision}, pages 20144--20154, 2023.

\bibitem{wu2023towards}
Haoning Wu, Erli Zhang, Liang Liao, Chaofeng Chen, Jingwen Hou, Annan Wang, Wenxiu Sun, Qiong Yan, and Weisi Lin.
\newblock Towards explainable in-the-wild video quality assessment: a database and a language-prompted approach.
\newblock In {\em Proceedings of the 31st ACM International Conference on Multimedia}, pages 1045--1054, 2023.

\bibitem{benjamin2018work}
Walter Benjamin.
\newblock The work of art in the age of mechanical reproduction.
\newblock In {\em A museum studies approach to heritage}, pages 226--243. Routledge, 2018.

\bibitem{cherti2023reproducible}
Mehdi Cherti, Romain Beaumont, Ross Wightman, Mitchell Wortsman, Gabriel Ilharco, Cade Gordon, Christoph Schuhmann, Ludwig Schmidt, and Jenia Jitsev.
\newblock Reproducible scaling laws for contrastive language-image learning.
\newblock In {\em Proceedings of the IEEE/CVF Conference on Computer Vision and Pattern Recognition}, pages 2818--2829, 2023.

\bibitem{sun2023eva}
Quan Sun, Yuxin Fang, Ledell Wu, Xinlong Wang, and Yue Cao.
\newblock Eva-clip: Improved training techniques for clip at scale.
\newblock {\em arXiv preprint arXiv:2303.15389}, 2023.

\bibitem{dosovitskiy2021image}
Alexey Dosovitskiy, Lucas Beyer, Alexander Kolesnikov, Dirk Weissenborn, Xiaohua Zhai, Thomas Unterthiner, Mostafa Dehghani, Matthias Minderer, Georg Heigold, Sylvain Gelly, et~al.
\newblock An image is worth 16x16 words: Transformers for image recognition at scale.
\newblock In {\em International Conference on Learning Representations}, 2021.

\bibitem{liu2022convnet}
Zhuang Liu, Hanzi Mao, Chao-Yuan Wu, Christoph Feichtenhofer, Trevor Darrell, and Saining Xie.
\newblock A convnet for the 2020s.
\newblock In {\em Proceedings of the IEEE/CVF conference on computer vision and pattern recognition}, pages 11976--11986, 2022.

\bibitem{lisncodalab2024imagecompetition}
Chunyi Li, Tengchuan Kou, Yixuan Gao, Yuqin Cao, Wei Sun, Zicheng Zhang, Yingjie Zhou, Zhichao Zhang, Weixia Zhang, Haoning Wu, et~al.
\newblock Aigiqa-20k: A large database for ai-generated image quality assessment.
\newblock {\em arXiv preprint arXiv:2404.03407}, 2024.

\bibitem{lisncodalab2024videocompetition}
Tengchuan Kou, Xiaohong Liu, Zicheng Zhang, Chunyi Li, Haoning Wu, Xiongkuo Min, Guangtao Zhai, and Ning Liu.
\newblock Subjective-aligned dateset and metric for text-to-video quality assessment.
\newblock {\em arXiv preprint arXiv:2403.11956}, 2024.

\bibitem{moorthy2011blind}
Anush~Krishna Moorthy and Alan~Conrad Bovik.
\newblock Blind image quality assessment: From natural scene statistics to perceptual quality.
\newblock {\em IEEE transactions on Image Processing}, 20(12):3350--3364, 2011.

\bibitem{saad2012blind}
Michele~A Saad, Alan~C Bovik, and Christophe Charrier.
\newblock Blind image quality assessment: A natural scene statistics approach in the dct domain.
\newblock {\em IEEE transactions on Image Processing}, 21(8):3339--3352, 2012.

\bibitem{ye2012unsupervised}
Peng Ye, Jayant Kumar, Le~Kang, and David Doermann.
\newblock Unsupervised feature learning framework for no-reference image quality assessment.
\newblock In {\em IEEE conference on computer vision and pattern recognition}, pages 1098--1105. IEEE, 2012.

\bibitem{ma2017dipiq}
Kede Ma, Wentao Liu, Tongliang Liu, Zhou Wang, and Dacheng Tao.
\newblock dipiq: Blind image quality assessment by learning-to-rank discriminable image pairs.
\newblock {\em IEEE Transactions on Image Processing}, 26(8):3951--3964, 2017.

\bibitem{kang2014convolutional}
Le~Kang, Peng Ye, Yi~Li, and David Doermann.
\newblock Convolutional neural networks for no-reference image quality assessment.
\newblock In {\em Proceedings of the IEEE conference on computer vision and pattern recognition}, pages 1733--1740, 2014.

\bibitem{bosse2017deep}
Sebastian Bosse, Dominique Maniry, Klaus-Robert M{\"u}ller, Thomas Wiegand, and Wojciech Samek.
\newblock Deep neural networks for no-reference and full-reference image quality assessment.
\newblock {\em IEEE Transactions on image processing}, 27(1):206--219, 2017.

\bibitem{ma2017end}
Kede Ma, Wentao Liu, Kai Zhang, Zhengfang Duanmu, Zhou Wang, and Wangmeng Zuo.
\newblock End-to-end blind image quality assessment using deep neural networks.
\newblock {\em IEEE Transactions on Image Processing}, 27(3):1202--1213, 2017.

\bibitem{bianco2018use}
Simone Bianco, Luigi Celona, Paolo Napoletano, and Raimondo Schettini.
\newblock On the use of deep learning for blind image quality assessment.
\newblock {\em Signal, Image and Video Processing}, 12:355--362, 2018.

\bibitem{wu2023assessor360}
Tianhe Wu, Shuwei Shi, Haoming Cai, Mingdeng Cao, Jing Xiao, Yinqiang Zheng, and Yujiu Yang.
\newblock Assessor360: Multi-sequence network for blind omnidirectional image quality assessment.
\newblock {\em Advances in Neural Information Processing Systems}, 36, 2023.

\bibitem{wang2021rich}
Yilin Wang, Junjie Ke, Hossein Talebi, Joong~Gon Yim, Neil Birkbeck, Balu Adsumilli, Peyman Milanfar, and Feng Yang.
\newblock Rich features for perceptual quality assessment of ugc videos.
\newblock In {\em Proceedings of the IEEE/CVF Conference on Computer Vision and Pattern Recognition}, pages 13435--13444, 2021.

\bibitem{goodfellow2014generative}
Ian Goodfellow, Jean Pouget-Abadie, Mehdi Mirza, Bing Xu, David Warde-Farley, Sherjil Ozair, Aaron Courville, and Yoshua Bengio.
\newblock Generative adversarial nets.
\newblock {\em Advances in neural information processing systems}, 27, 2014.

\bibitem{kingma2014auto}
Diederik~P Kingma and Max Welling.
\newblock Auto-encoding variational bayes.
\newblock In {\em International Conference on Learning Representations}, 2014.

\bibitem{ho2020denoising}
Jonathan Ho, Ajay Jain, and Pieter Abbeel.
\newblock Denoising diffusion probabilistic models.
\newblock {\em Advances in neural information processing systems}, 33:6840--6851, 2020.

\bibitem{song2021score}
Yang Song, Jascha Sohl-Dickstein, Diederik~P Kingma, Abhishek Kumar, Stefano Ermon, and Ben Poole.
\newblock Score-based generative modeling through stochastic differential equations.
\newblock In {\em International Conference on Learning Representations}, 2021.

\bibitem{rombach2022high}
Robin Rombach, Andreas Blattmann, Dominik Lorenz, Patrick Esser, and Bj{\"o}rn Ommer.
\newblock High-resolution image synthesis with latent diffusion models.
\newblock In {\em Proceedings of the IEEE/CVF conference on computer vision and pattern recognition}, pages 10684--10695, 2022.

\bibitem{podell2024sdxl}
Dustin Podell, Zion English, Kyle Lacey, Andreas Blattmann, Tim Dockhorn, Jonas M{\"u}ller, Joe Penna, and Robin Rombach.
\newblock Sdxl: Improving latent diffusion models for high-resolution image synthesis.
\newblock In {\em The Twelfth International Conference on Learning Representations}, 2024.

\bibitem{blattmann2023stable}
Andreas Blattmann, Tim Dockhorn, Sumith Kulal, Daniel Mendelevitch, Maciej Kilian, Dominik Lorenz, Yam Levi, Zion English, Vikram Voleti, Adam Letts, et~al.
\newblock Stable video diffusion: Scaling latent video diffusion models to large datasets.
\newblock {\em arXiv preprint arXiv:2311.15127}, 2023.

\bibitem{peebles2023scalable}
William Peebles and Saining Xie.
\newblock Scalable diffusion models with transformers.
\newblock In {\em Proceedings of the IEEE/CVF International Conference on Computer Vision}, pages 4195--4205, 2023.

\bibitem{vaswani2017attention}
Ashish Vaswani, Noam Shazeer, Niki Parmar, Jakob Uszkoreit, Llion Jones, Aidan~N Gomez, {\L}ukasz Kaiser, and Illia Polosukhin.
\newblock Attention is all you need.
\newblock {\em Advances in neural information processing systems}, 30, 2017.

\bibitem{openai2024sora}
OpenAI.
\newblock Sora: A review on background, technology, limitations, and opportunities of large vision models.
\newblock \url{https://openai.com/research/video-generation-models-as-world-simulators}, 2024.

\bibitem{ruiz2023dreambooth}
Nataniel Ruiz, Yuanzhen Li, Varun Jampani, Yael Pritch, Michael Rubinstein, and Kfir Aberman.
\newblock Dreambooth: Fine tuning text-to-image diffusion models for subject-driven generation.
\newblock In {\em Proceedings of the IEEE/CVF Conference on Computer Vision and Pattern Recognition}, pages 22500--22510, 2023.

\bibitem{ramesh2022hierarchical}
Aditya Ramesh, Prafulla Dhariwal, Alex Nichol, Casey Chu, and Mark Chen.
\newblock Hierarchical text-conditional image generation with clip latents.
\newblock {\em arXiv preprint arXiv:2204.06125}, 2022.

\bibitem{betker2023improving}
James Betker, Gabriel Goh, Li~Jing, Tim Brooks, Jianfeng Wang, Linjie Li, Long Ouyang, Juntang Zhuang, Joyce Lee, Yufei Guo, et~al.
\newblock Improving image generation with better captions.
\newblock {\em Computer Science. https://cdn. openai. com/papers/dall-e-3. pdf}, 2(3):8, 2023.

\bibitem{li2023agiqa}
Chunyi Li, Zicheng Zhang, Haoning Wu, Wei Sun, Xiongkuo Min, Xiaohong Liu, Guangtao Zhai, and Weisi Lin.
\newblock Agiqa-3k: An open database for ai-generated image quality assessment.
\newblock {\em IEEE Transactions on Circuits and Systems for Video Technology}, 2023.

\bibitem{mu2022slip}
Norman Mu, Alexander Kirillov, David Wagner, and Saining Xie.
\newblock Slip: Self-supervision meets language-image pre-training.
\newblock In {\em European conference on computer vision}, pages 529--544. Springer, 2022.

\bibitem{li2022blip}
Junnan Li, Dongxu Li, Caiming Xiong, and Steven Hoi.
\newblock Blip: Bootstrapping language-image pre-training for unified vision-language understanding and generation.
\newblock In {\em International conference on machine learning}, pages 12888--12900. PMLR, 2022.

\bibitem{li2023blip}
Junnan Li, Dongxu Li, Silvio Savarese, and Steven Hoi.
\newblock Blip-2: Bootstrapping language-image pre-training with frozen image encoders and large language models.
\newblock In {\em International conference on machine learning}, pages 19730--19742. PMLR, 2023.

\bibitem{yang2023attentive}
Yifan Yang, Weiquan Huang, Yixuan Wei, Houwen Peng, Xinyang Jiang, Huiqiang Jiang, Fangyun Wei, Yin Wang, Han Hu, Lili Qiu, et~al.
\newblock Attentive mask clip.
\newblock In {\em Proceedings of the IEEE/CVF International Conference on Computer Vision}, pages 2771--2781, 2023.

\bibitem{fang2023data}
Alex Fang, Albin~Madappally Jose, Amit Jain, Ludwig Schmidt, Alexander Toshev, and Vaishaal Shankar.
\newblock Data filtering networks.
\newblock {\em arXiv preprint arXiv:2309.17425}, 2023.

\bibitem{schuhmann2022laion}
Christoph Schuhmann, Romain Beaumont, Richard Vencu, Cade Gordon, Ross Wightman, Mehdi Cherti, Theo Coombes, Aarush Katta, Clayton Mullis, Mitchell Wortsman, et~al.
\newblock Laion-5b: An open large-scale dataset for training next generation image-text models.
\newblock {\em Advances in Neural Information Processing Systems}, 35:25278--25294, 2022.

\bibitem{gadre2024datacomp}
Samir~Yitzhak Gadre, Gabriel Ilharco, Alex Fang, Jonathan Hayase, Georgios Smyrnis, Thao Nguyen, Ryan Marten, Mitchell Wortsman, Dhruba Ghosh, Jieyu Zhang, et~al.
\newblock Datacomp: In search of the next generation of multimodal datasets.
\newblock {\em Advances in Neural Information Processing Systems}, 36, 2024.

\bibitem{chen2022pali}
Xi~Chen, Xiao Wang, Soravit Changpinyo, AJ~Piergiovanni, Piotr Padlewski, Daniel Salz, Sebastian Goodman, Adam Grycner, Basil Mustafa, Lucas Beyer, et~al.
\newblock Pali: A jointly-scaled multilingual language-image model.
\newblock {\em arXiv preprint arXiv:2209.06794}, 2022.

\bibitem{cai2023efficientvit}
Han Cai, Junyan Li, Muyan Hu, Chuang Gan, and Song Han.
\newblock Efficientvit: Lightweight multi-scale attention for high-resolution dense prediction.
\newblock In {\em Proceedings of the IEEE/CVF International Conference on Computer Vision}, pages 17302--17313, 2023.

\bibitem{fang2023eva}
Yuxin Fang, Quan Sun, Xinggang Wang, Tiejun Huang, Xinlong Wang, and Yue Cao.
\newblock Eva-02: A visual representation for neon genesis.
\newblock {\em arXiv preprint arXiv:2303.11331}, 2023.

\bibitem{loshchilov2019decoupled}
Ilya Loshchilov and Frank Hutter.
\newblock Decoupled weight decay regularization.
\newblock In {\em International Conference on Learning Representations}, 2019.

\bibitem{he2016deep}
Kaiming He, Xiangyu Zhang, Shaoqing Ren, and Jian Sun.
\newblock Deep residual learning for image recognition.
\newblock In {\em Proceedings of the IEEE conference on computer vision and pattern recognition}, pages 770--778, 2016.

\bibitem{dosovitskiy2020image}
Alexey Dosovitskiy, Lucas Beyer, Alexander Kolesnikov, Dirk Weissenborn, Xiaohua Zhai, Thomas Unterthiner, Mostafa Dehghani, Matthias Minderer, Georg Heigold, Sylvain Gelly, et~al.
\newblock An image is worth 16x16 words: Transformers for image recognition at scale.
\newblock {\em arXiv preprint arXiv:2010.11929}, 2020.

\bibitem{zhang2024long}
Beichen Zhang, Pan Zhang, Xiaoyi Dong, Yuhang Zang, and Jiaqi Wang.
\newblock Long-clip: Unlocking the long-text capability of clip.
\newblock {\em arXiv preprint arXiv:2403.15378}, 2024.

\bibitem{fang2021stochastic}
Xi~Fang, Jiancheng Yang, and Bingbing Ni.
\newblock Stochastic label refinery: Toward better target label distribution.
\newblock In {\em 2020 25th International Conference on Pattern Recognition (ICPR)}, pages 9115--9121. IEEE, 2021.

\bibitem{liu2024challenge}
Xiaohong Liu, Xiongkuo Min, Guangtao Zhai, Chunyi Li, Tengchuan Kou, Wei Sun, Haoning Wu, Yixuan Gao, Yuqin Cao, Zicheng Zhang, Xiele Wu, and Radu Timofte.
\newblock Ntire 2024 quality assessment of ai-generated content challenge.
\newblock In {\em CVPR Workshop}, 2024.

\end{thebibliography}
}


\end{document}